\documentclass[10pt,twocolumn,letterpaper]{article}

\usepackage{iccv}

\usepackage{times}
\usepackage{epsfig}
\usepackage{graphicx}
\usepackage{amsmath}
\usepackage{amssymb}
\usepackage{multirow}
\usepackage{xspace}
\usepackage{comment}
\usepackage[british,UKenglish,USenglish,english,american]{babel}
\newcommand{\tabincell}[2]{\begin{tabular}{@{}#1@{}}#2\end{tabular}}

\usepackage{amssymb}
\usepackage{pifont}

\newcommand{\cmark}{\ding{51}}%
\newcommand{\xmark}{\ding{55}}%

\usepackage[pagebackref=true,breaklinks=true,letterpaper=true,colorlinks,bookmarks=false]{hyperref}

\iccvfinalcopy 

\def\iccvPaperID{971} 

\ificcvfinal\fi

\begin{document}

\title{Compositional Human Pose Regression}


{
\author{Xiao Sun$^1$, Jiaxiang Shang$^1$, Shuang Liang$^2$\thanks{Corresponding author.}, Yichen Wei$^1$ \\
$^1$Microsoft Research, $^2$ Tongji University\\
	{\tt\small  \{xias, yichenw\}@microsoft.com, jiaxiang.shang@gmail.com, shuangliang@tongji.edu.cn}
}
}

\maketitle

\begin{abstract}
Regression based methods are not performing as well as detection based methods for human pose estimation. A central problem is that the structural information in the pose is not well exploited in the previous regression methods. In this work, we propose a structure-aware regression approach. It adopts a reparameterized pose representation using bones instead of joints. It exploits the joint connection structure to define a compositional loss function that encodes the long range interactions in the pose. It is simple, effective, and general for both 2D and 3D pose estimation in a unified setting. Comprehensive evaluation validates the effectiveness of our approach. It significantly advances the state-of-the-art on Human3.6M~\cite{ionescu2014human3} and is competitive with state-of-the-art results on MPII~\cite{andriluka20142d}.
\end{abstract}

\section{Introduction}
\label{sec.introduction}

Human pose estimation has been extensively studied for both 3D~\cite{ionescu2014human3} and 2D~\cite{andriluka20142d}. Recently, deep convolutional neutral networks (CNNs) have achieved significant progresses. 

Existing approaches fall into two categories: detection based and regression based. Detection based methods generate a likelihood heat map for each joint and locate the joint as the point with the maximum value in the map. These heat maps are usually noisy and multi-mode. The ambiguity is reduced by exploiting the dependence between the joints in various ways. A prevalent family of state-of-the-art methods~\cite{chu2017multi, carreira2016human, newell2016stacked, bulat2016human, wei2016convolutional, insafutdinov2016deepercut} adopt a multi-stage architecture, where the output of the previous stage is used as input to enhance the learning of the next stage. These methods are dominant for 2D pose estimation~\cite{mpiiwebpage}. However, they do not easily generalize to 3D pose estimation, because the 3D heat maps are too demanding for memory and computation.

Regression based methods directly map the input image to the output joints. They  directly target at the task and they are general for both 3D and 2D pose estimation. Nevertheless, they are not performing as well as detection based methods. As an evidence, only one method~\cite{carreira2016human} in the 2D pose benchmark ~\cite{mpiiwebpage} is regression based. While they are widely used for 3D pose estimation ~\cite{zhou2016deep,moreno20163d,mehta2016monocular,tekin2016direct,li2015maximum,tekin2016structured,park20163d}, the performance is not satisfactory. A central problem is that they simply minimize the per-joint location errors \emph{independently} but ignore the internal structures of the pose. In other words, joint dependence is not well exploited.

In this work, we propose a structure-aware approach, called \emph{compositional pose regression}. It is based on two ideas. First, it uses bones instead of joints as pose representation, because the bones are more primitive, more stable, and easier to learn than joints. Second, it exploits the joint connection structure to define a \emph{compositional loss function} that encodes long range interactions between the bones.

The approach is simple, effective and efficient. It only re-parameterizes the pose representation, which is the network output, and enhances the loss function, which relates the output to ground truth. It does not alter other algorithm design choices and is compatible with such choices, such as network architecture. It can be easily adapted into any existing regression approaches with little overhead for memory and computation, in both training and inference.

The approach is general and can be used for both 3D and 2D pose regression, indistinguishably. Moreover, 2D and 3D data can be easily mixed simultaneously in the training. \emph{For the first time, it is shown that such directly mixed learning is effective}. This property makes our approach different from all existing ones that target at either 3D or 2D task.

The effectiveness of our approach is validated by comprehensive evaluation with a few new metrics, rigorous ablation study and comparison with state-of-the-art on both 3D and 2D benchmarks. Specifically, it advances the state-of-the-art on 3D Human3.6M dataset~\cite{ionescu2014human3} by a large margin and achieves a record of $59.1$ mm average joint error, about $12\%$ relatively better that state-of-the-art. On 2D MPII dataset~\cite{andriluka20142d, mpiiwebpage}, it achieves $86.4\%$ (PCKh 0.5). It is the best-performing regression based method and on bar with the state-of-the-art detection based methods. As a by-product, our approach generates high quality 3D poses for in the wild images, indicating the potential of our approach for transfer learning of 3D pose estimation in the wild.

\section{Related Work}
\label{sec.related_work}

Human pose estimation has been extensively studied for years. A complete review is beyond the scope of this work. We refer the readers to ~\cite{moeslund2001survey,sarafianos20163d} for a detailed survey.

The previous works are reviewed from two perspectives related to this work. First is how to exploit the joint dependency for 3D and 2D pose estimation. Second is how to exploit ``in the wild'' 2D data for 3D pose estimation.

\textbf{3D Pose Estimation} Some methods use two separate steps. They first perform 2D joint prediction and then reconstruct the 3D pose via optimization or search. There is no end-to-end learning. Zhou et al.~\cite{zhou2016sparseness} combines uncertainty maps of the 2D joints location and a sparsity-driven 3D geometric prior to infer the 3D joint location via an EM algorithm. Chen et al.~\cite{chen20163d} searches a large 3D pose library and uses the estimated 2D pose as query. Bogo et al.~\cite{bogo2016keep} fit a recently published statistical body shape model ~\cite{loper2015smpl} to the 2D joints. Jahangiri et al.~\cite{jahangiri2017generating} generates multiple hypotheses from 2D joints using a novel generative model.

Some methods implicitly learn the pose structure from data. Tekin et al.~\cite{tekin2016structured} represents the 3D pose with an over-complete dictionary. A high-dimensional latent pose representation is learned to account for joint dependencies. Pavlakos et al.~\cite{pavlakos2016coarse} extends the Hourglass~\cite{newell2016stacked} framework from 2D to 3D. A coarse-to-fine approach is used to address the large dimensionality increase. Li et al.~\cite{li2015maximum} uses an image-pose embedding sub-network to regularize the 3D pose prediction.



Above works do not use prior knowledge in 3D model. Such prior knowledge is firstly used in ~\cite{zhou2016deep,zhou2016model} by embedding a kinematic model layer into deep neutral networks and estimating model parameters instead of joints. The geometric structure is better preserved. Yet, the kinematic model parameterization is highly nonlinear and its optimization in deep networks is hard. Also, the methods are limited for a fully specified kinematic model (fixed bone length, known scale). They do not generalize to 2D pose estimation, where a good 2D kinematic model does not exist.

\textbf{2D Pose Estimation} Before the deep learning era, many methods use graphical models to represent the structures in the joints. Pictorial structure model~\cite{felzenszwalb2005pictorial} is one of the earliest. There is a lot of extensions~\cite{johnson2011learning, yang2011articulated, pishchulin2013poselet, pedersoli2015coarse, yang2013articulated, lindner2015robust, chen2014articulated}. Pose estimation is formulated as inference problems on the graph. A common drawback is that the inference is usually complex, slow, and hard to integrate with deep networks.


Recently, the graphical models have been integrated into deep networks in various ways. Tompson et al.~\cite{tompson2014joint} firstly combine a convolutional network with a graphical model for human pose estimation. Ouyang et al. ~\cite{ouyang2013joint} joints feature extraction, part deformation handling, occlusion handling and classification all into deep learning framework. Chu et al.~\cite{chu2016structured} introduce a geometrical transform kernels in CNN framework that can pass informations between different joint heat maps. Both features and their relationships are jointly learned in a end-to-end learning system. Yang et al.~\cite{yang2016end} combine deep CNNs with the expressive deformable mixture of parts to regularize the output.



Another category of methods use a multi-stage architecture~\cite{chu2017multi, carreira2016human, newell2016stacked, bulat2016human, wei2016convolutional, insafutdinov2016deepercut, gkioxari2016chained}. The results of the previous stage are used as inputs to enhance or regularize the learning of the next stage.  Newell et al.~\cite{newell2016stacked} introduce an Stacked Hourglass architecture that better capture the various spatial relationships associated with the body. Chu et al.~\cite{chu2017multi} further extend ~\cite{newell2016stacked} with a multi-context attention mechanism. Bulat et al.~\cite{bulat2016human} propose a detection-followed-by-regression CNN cascade. Wei et al.~\cite{wei2016convolutional} design a sequential architecture composed of convolutional networks that directly operate on belief maps from previous stages. Gkioxari et al.~\cite{gkioxari2016chained} predict joint heat maps sequentially and conditionally according to their difficulties. All such methods learn the joint dependency from data, implicitly.


Different to all above 3D and 2D methods, our approach explicitly exploits the joint connection structure in the pose. It does not make further assumptions and does not involve complex algorithm design. It only changes the pose representation and enhances the loss function. It is simple, effective, and can be combined with existing techniques.

\textbf{Leveraging in the wild 2D data for 3D pose estimation} 3D pose capturing is difficult. The largest 3D human pose dataset Human3.6M~\cite{ionescu2014human3} is still limited in that the subjects, the environment, and the poses have limited complexity and variations. Models trained on such data do not generalize well to other domains, such as in the wild images.

In contrast, in the wild images and 2D pose annotation are abundant. Many works leverage the 2D data for 3D pose estimation. Most of them consist of two separate steps.

Some methods firstly generate the 2D pose results (joint locations or heat maps) and then use them as input for recovering the 3D pose. The information in the 2D images is discarded in the second step. Bogo et al.~\cite{bogo2016keep} first use DeepCut ~\cite{pishchulin2016deepcut} to generate 2D joint location, then fit with a 3D body shape model. Moreno et al.~\cite{moreno20163d} use CPM ~\cite{wei2016convolutional} to detect 2D position of human joints, and then use these observations to infer 3D pose via distance matrix regression. Zhou et al.~\cite{zhou2017monocap} use Hourglass~\cite{newell2016stacked} to generate 2D joint heat maps and then coupled with a geometric prior and Jahangiri et al.~\cite{jahangiri2017generating} also use Hourglass to predict 2D joint heat maps and then infer multiple 3D hypotheses from them. Wu et al.~\cite{wu2016single} propose 3D interpreter network that sequentially estimates 2D keypoint heat maps and 3D object structure. 

Some methods firstly train the deep network model on 2D data and fine-tune the model on 3D data. The information in 2D data is partially retained by the pre-training, but not fully exploited as the second fine-tuning step cannot use 2D data. Pavlakos et al.~\cite{pavlakos2016coarse} extends Hourglass~\cite{newell2016stacked} model for 3D volumetric prediction. 2D heat maps are used as intermediate supervision. Tome et al. ~\cite{tome2017lifting} extends CPM ~\cite{wei2016convolutional} to 3D by adding a probabilistic 3D pose model to the CPM.

Some methods train both 2D and 3D pose networks simultaneously by sharing intermediate CNN features ~\cite{mehta2016monocular, park20163d}. Yet, they use separate networks for 2D and 3D tasks.



Unlike the above methods, our approach treats the 2D and 3D data in the same way and combine them in a unified training framework. The abundant information in the 2D data is fully exploited during training. As a result, our method achieves strong performance on both 3D and 2D benchmarks. As a by-product, it generates plausible and convincing 3D pose results for in the wild images.

Some methods use synthetic datasets which are generated from deforming a human template model with known ground truth ~\cite{chen2016synthesizing, rogez2016mocap}. These methods are complementary to the others as they focus on data augmentation.

\begin{figure*}[t]
  \centering
   \includegraphics [width=0.22\linewidth] {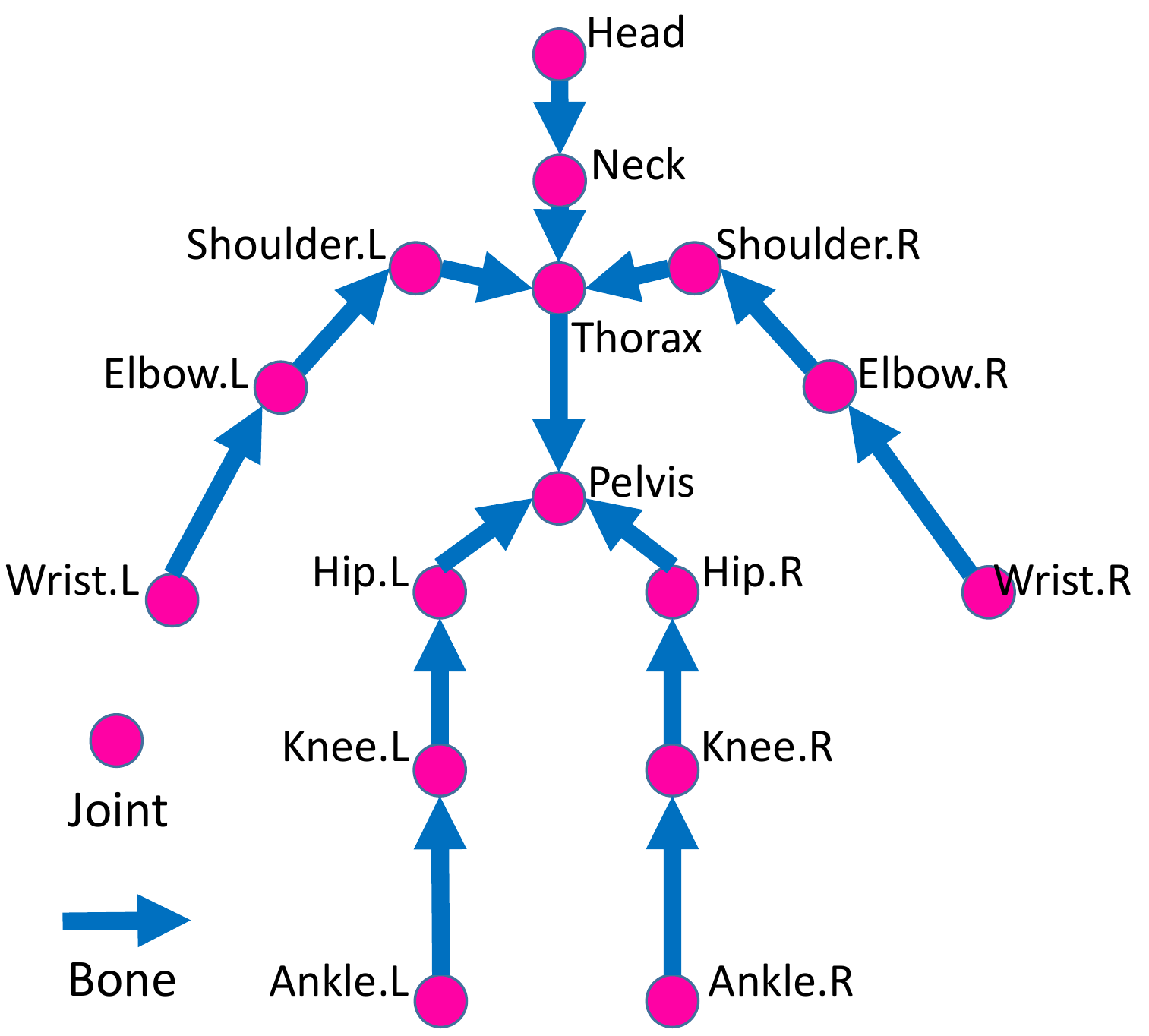}
   \includegraphics [width=0.33\linewidth] {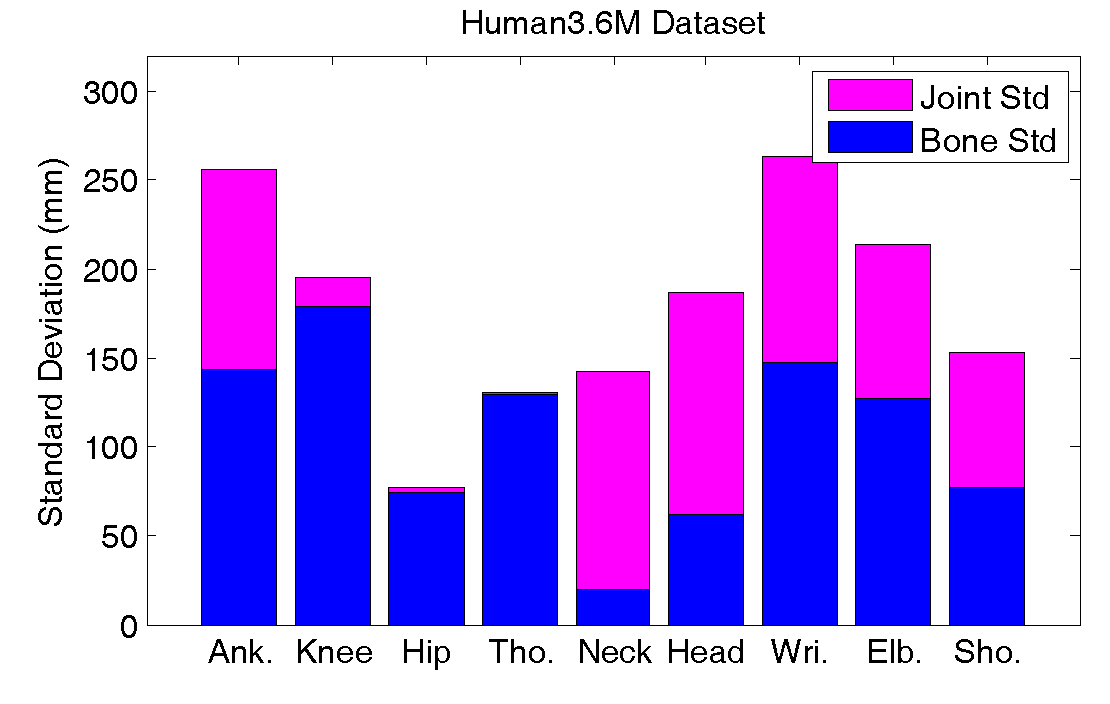}
   \includegraphics [width=0.33\linewidth] {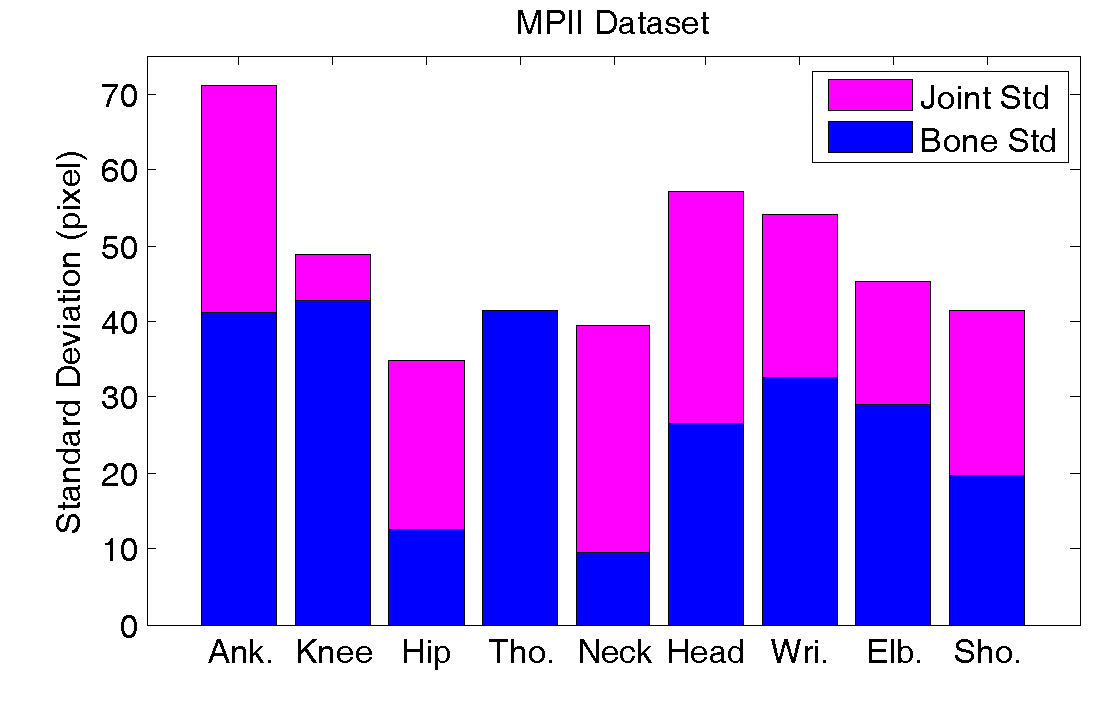}
    \caption{Left: a human pose is represented as either joints $\mathcal{J}$ or bones $\mathcal{B}$. Middle/Right: standard deviations of bones and joints for the 3D Human3.6M dataset~\cite{ionescu2014human3} and 2D MPII dataset~\cite{andriluka20142d}.}
    \label{fig.bone_representation}
\end{figure*}

\section{Compositional Pose Regression}
\label{sec.compositional_pose_regression}

Given an image of a person, the pose estimation problem is to obtain the 2D (or 3D) position of all the $K$ joints, $\mathcal{J}=\{\mathbf{J}_k|k=1,...,K\}$. Typically, the coordinate unit is \emph{pixel} for 2D and \emph{millimeter (mm)} for 3D.

Without loss of generality, the joints are defined with respect to a \emph{constant origin point} in the image coordinate system. For convenience, let the origin be $\mathbf{J}_0$. Specifically, for 2D pose estimation, it is the top-left point of the image. For 3D pose estimation, it is the ground truth pelvis joint~\cite{zhou2016deep,park20163d}.

For regression learning, normalization is necessary to compensate for the differences in  magnitude of the variables. We use the standard normalization by subtraction of mean and division of standard deviation. For a variable \emph{var}, it is normalized as

\begin{equation}
\tilde{var}=N(var)=\frac{var-mean(var^{gt})}{std(var^{gt})}.
\label{eq.normalization}
\end{equation}

The inverse function for \emph{unnormalization} is
\begin{equation}
var=N^{-1}(\tilde{var}) = \tilde{var}\cdot std(var^{gt})+mean(var^{gt}).
\label{eq.unnormalization}
\end{equation}

Note that both $mean(*)$ and $std(*)$ are constants and calculated from the ground truth training samples. The predicted output from the network is assumed already normalized. Both functions $N(*)$ and $N^{-1}(*)$ are parameter free and embedded in the network. For notation simplicity, we use $\tilde{var}$ for $N(var)$.

\subsection{Direct Joint Regression: A Baseline}
\label{sec.direct_joint_regression}

Most previous regression based methods~\cite{carreira2016human,zhou2016deep,park20163d,tekin2016structured,tekin2016direct} directly minimize the squared difference of the predicted and ground truth joints. In experiments, we found that the absolute difference ($L1$ norm) performs better. In our \emph{direct joint regression} baseline, the joint loss is

\begin{equation}
L(\mathcal{J}) = \sum_{k=1}^K ||\tilde{\mathbf{J}}_k - \tilde{\mathbf{J}}_k^{gt}||_1.
\label{eq.direct_joint_loss}
\end{equation}

Note that both the prediction and ground truth are normalized.

There is a clear drawback in loss Eq.\eqref{eq.direct_joint_loss}. The joints are \emph{independently} estimated. The joint correlation, or the internal structure in the pose, is not well exploited. For example, certain geometric constraints (\eg, bone length is fixed) are not satisfied.

Previous works only evaluate the joint location accuracy. This is also limited because the internal structures in the pose are not well evaluated.

\subsection{A Bone Based Representation}
\label{sec.bon_representation}

We show that a simple \emph{reparameterization} of the pose is effective to address the above issues. As shown in Figure~\ref{fig.bone_representation}(left), a pose is structured as a tree. Without loss of generality, let pelvis be the the root joint $\mathbf{J}_1$ and tree edges be directed from the root to the end joints such as wrists and ankles. Let the function $parent(k)$ return the index of parent joint for $k^{th}$ joint. For notation consistency, let the parent of the root joint $\mathbf{J}_1$ be the origin $\mathbf{J}_0$, \ie, $parent(1)=0$.

Now, for $k^{th}$ joint, we define its associated bone as a directed vector pointing from it to its parent,

\begin{equation}
\mathbf{B}_k = \mathbf{J}_{parent(k)} - \mathbf{J}_k.
\label{eq.bone_definition}
\end{equation}

The joints $\mathcal{J}$ are defined in the global coordinate system. In contrast, bones $\mathcal{B}=\{\mathbf{B}_k|k=1,...,K\}$ are more primitive and defined in the local coordinate systems. Representing the pose using bones brings several benefits.

\textbf{Stability} Bones are more stable than joints and easier to learn. Figure~\ref{fig.bone_representation} (middle and right) shows that the standard deviation of bones is much smaller than that of their corresponding joints, especially for parts (ankle, wrist, head) far away from the root pelvis, in both 3D (Human 3.6M~\cite{ionescu2014human3}) and 2D datasets (MPII~\cite{andriluka20142d}).

\textbf{Geometric convenience} Bones can encode the geometric structure and express the geometric constraints more easily than joints. For example, constraint of \emph{``bone length is fixed''} involves one bone but two joints. Constraint of \emph{``joint rotation angle is in limited range''} involves two bones but three joints. Such observations motivate us to propose new evaluation metrics for \emph{geometric validity}, as elaborated in Section~\ref{sec.exp}. Experiments show that bone based representation is better than joint based representation on such metrics.

\textbf{Application convenience} Many pose-driven applications only need the local bones instead of the global joints. For example, the local and relative ``elbow to wrist'' motion can sufficiently represent a ``pointing'' gesture that would be useful for certain human computer interaction scenarios.

\subsection{Compositional Loss Function}
\label{sec.compositional_loss}

Similar to the joint loss in Eq.~\eqref{eq.direct_joint_loss}, bones can be learnt by minimizing the bone loss function

\begin{equation}
L(\mathcal{B}) = \sum_{k=1}^K ||\tilde{\mathbf{B}}_k - \tilde{\mathbf{B}}_k^{gt}||_1.
\label{eq.direct_bone_loss}
\end{equation}

However, there is a clear drawback in this loss. As the bones are local and \emph{independently} estimated in Eq.~\eqref{eq.direct_bone_loss}, the errors in the individual bone predictions would propagate along the skeleton and accumulate into large errors for joints at the far end. For example, in order to predict $\mathbf{J}_{wrist}$, we need to concatenate $\mathbf{B}_{wrist}$, $\mathbf{B}_{elbow}$,...,$\mathbf{B}_{pelvis}$. Errors in these bones will accumulate and affect the accuracy of $\mathbf{J}_{wrist}$ in a random manner.

To address the problem, long range objectives should be considered in the loss. Long range errors should be balanced over the intermediate bones. In this way, bones are \emph{jointly} optimized. Specifically, let $\mathbf{J}_u$ and $\mathbf{J}_v$ be two arbitrary joints. Suppose the path from $\mathbf{J}_u$ to $\mathbf{J}_v$ along the skeleton tree has $M$ joints. Let the function $I(m)$ return the index of the $m^{th}$ joint on the path, \eg, $I(1)=u$, $I(M)=v$. Note that $M$ and $I(*)$ are constants but depend on $u$ and $v$. Such dependence is omitted in the notations for clarity.

The long range, relative joint position $\Delta\mathbf{J}_{u,v}$ is the summation of the bones along the path, as

\begin{equation}
\begin{split}
\Delta\mathbf{J}_{u,v} &  = \sum_{m = 1}^{M-1} \mathbf{J}_{I(m+1)} - \mathbf{J}_{I(m)}\\
   & =\sum_{m=1}^{M-1}sgn(parent(I(m)),I(m+1))\cdot N^{-1}(\tilde{\mathbf{B}}_{I(m)}).\\
\end{split}
\label{eq.delta_joint_composition}
\end{equation}

The function $sgn(*,*)$ indicates whether the bone $\mathbf{B}_{I(m)}$ direction is along the path direction. It returns $1$ when $parent(I(m))=I(m+1)$ and $-1$ otherwise. 
Note that the network predicted bone $\tilde{\mathbf{B}}(*)$ is normalized, as in Eq.~\eqref{eq.delta_joint_composition}. It is unnormalized via Eq.~\eqref{eq.unnormalization} before summation.

Eq.\eqref{eq.delta_joint_composition} is differentiable with respect to the bones. It is efficient and has no free parameters. It is implemented as a special \emph{compositional} layer in the neutral networks.

The ground truth relative position is
\begin{equation}
\Delta\mathbf{J}_{u,v}^{gt} = \mathbf{J}_u^{gt} - \mathbf{J}_v^{gt}.
\end{equation}

Then, given a joint pair set $\mathcal{P}$, the \emph{compositional loss function} is defined as
\begin{equation}
L(\mathcal{B},\mathcal{P}) = \sum_{(u,v)\in\mathcal{P}}||\tilde{\Delta\mathbf{J}}_{u,v} - \tilde{\Delta\mathbf{J}}_{u,v}^{gt}||_1.
\label{eq.compositional_loss_function}
\end{equation}

In this way, every joint pair $(u,v)$ constrains the bones along the path from $u$ to $v$. Each bone is constrained by multiple paths given a large number of joint pairs. The errors are better balanced over the bones during learning.

The joint pair set $\mathcal{P}$ can be arbitrary. To validate the effectiveness of Eq.\eqref{eq.compositional_loss_function}, we test four variants:

\begin{itemize}
\item $\mathcal{P}_{joint}=\{(u, 0)|u = 1,...,K\}$. It only considers the global joint locations. It is similar to joint loss Eq.\eqref{eq.direct_joint_loss}.

\item $\mathcal{P}_{bone}=\{(u,parent(u))|u = 1,...,K\}$. It only considers the bones. It degenerates to the bone loss Eq.\eqref{eq.direct_bone_loss}.

\item $\mathcal{P}_{both}=\mathcal{P}_{joint}\bigcup\mathcal{P}_{bone}$. It combines the above two and verifies whether Eq.\eqref{eq.compositional_loss_function} is effective.

\item $\mathcal{P}_{all}=\{(u,v)|u<v,u,v=1,...,K\}$. It contains all joint pairs. The pose structure is fully exploited.
\end{itemize}

\section{Unified 2D and 3D Pose Regression}
\label{sec.2d_3d_train}

All the notations and equations in Section~\ref{sec.compositional_pose_regression} are applicable for both 3D and 2D pose estimation in the same way. The output pose dimension is either $3K$ or $2K$.

Training using mixed 3D and 2D data is straightforward. All the variables, such as joint $\mathbf{J}$, bone $\mathbf{B}$, and relative joint position $\Delta\mathbf{J}_{u,v}$, are decomposed into $xy$ part and $z$ part.

The loss functions can be similarly decomposed. For example, for \emph{compositional loss function} Eq.\eqref{eq.compositional_loss_function}, we have

\begin{equation}
L(\mathcal{B},\mathcal{P}) = L_{xy}(\mathcal{B},\mathcal{P}) + L_{z}(\mathcal{B},\mathcal{P}).
\label{eq.loss_2d_plus_3d}
\end{equation}

The $xy$ term $L_{xy}(*,*)$ is always valid for both 3D and 2D samples. The $z$ term $L_{z}(*,*)$ is only computed for 3D samples and set to $0$ for 2D samples. In the latter case, no gradient is back-propagated from $L_{z}(*,*)$.


Note that the $xy$ part and $z$ part variables have different dimensions. $xy$ is in image coordinate frame and the unit is in pixel. $z$ is in camera coordinate frame and the unit is metric (millimeters in our case). This is no problem. During training, they are appropriately normalized (Eq.\eqref{eq.compositional_loss_function}, Eq.\eqref{eq.normalization}) or unnormalized (Eq.\eqref{eq.delta_joint_composition}, Eq.\eqref{eq.unnormalization}). During inference, in order to recover the 3D metric coordinates, the $xy$ part is back-projected into camera space using known camera intrinsic parameters and a perspecitive projection model.

\textbf{Training} We use the state-of-the-art ResNet-50~\cite{he2016deep}. The model is pre-trained on ImageNet classification dataset~\cite{deng2009imagenet}. The last fully connected layer is then modified to output $3K$ (or $2K$) coordinates and the model is fine-tuned on our target task and data. The training is the same for all the tasks (3D, 2D, mixed). SGD is used for optimization. There are 25 epoches. The base learning rate is 0.03. It drops to 0.003 after 10 epoches and 0.0003 after another 10 epoches. Mini-batch size is 64. Two GPUs are used. Weight decay is 0.0002. Momentum is 0.9. Batch-normalization ~\cite{ioffe2015batch} is used. Implementation is in Caffe ~\cite{jia2014caffe}.

\textbf{Data Processing and Augmentation} The input image is normalized to $224\times224$. Data augmentation includes random translation($\pm2\%$ of the image size), scale($\pm25\%$), rotation($\pm30$ degrees) and flip. For MPII dataset, the training data are augmented by 20 times. For Human3.6M dataset, the training data are augmented by 4 times. For mixed 2D-3D task, each mini-batch consists of half 2D and half 3D samples, randomly sampled and shuffled.

\section{Experiments}
\label{sec.exp}

Our approach is evaluated on 3D and 2D human pose benchmarks. \emph{Human3.6M~\cite{ionescu2014human3}} is the largest 3D human pose benchmark. The dataset is captured in controlled environment. The image appearance of the subjects and the background is simple. Accurate 3D human joint locations are obtained from motion capture devices.


\emph{MPII ~\cite{andriluka20142d}} is the benchmark dataset for 2D human pose estimation. It includes about $25k$ images and $40k$ annotated 2D poses. $25k$ of them are for training and another $7k$ of the remaining are for testing. The images were collected from YouTube videos covering daily human activities with complex poses and image appearances.

\subsection{Comprehensive Evaluation Metrics}

For 3D human pose estimation, previous works~\cite{chen20163d, tome2017lifting, moreno20163d, zhou2017monocap, jahangiri2017generating, mehta2016monocular, pavlakos2016coarse, yasin2016dual, rogez2016mocap, bogo2016keep, zhou2016sparseness, tekin2016direct, zhou2016deep} use the mean per joint position error (MPJPE). We call this metric \emph{Joint Error}. Some works ~\cite{yasin2016dual, rogez2016mocap, chen20163d, bogo2016keep, moreno20163d, zhou2017monocap} firstly align the predicted 3D pose and ground truth 3D pose with a rigid transformation using \emph{Procrustes Analysis} ~\cite{gower1975generalized} and then compute MPJPE. We call this metric \emph{PA Joint Error}.

For 2D human pose estimation in MPII ~\cite{andriluka20142d}, Percentage of Correct Keypoints (PCK) metric is used for evaluation.

Above metrics only measures the accuracy of \emph{absolute} joint location. They do not fully reflect the accuracy of internal structures in the pose. We propose three additional metrics for a comprehensive evaluation.

The first metric is the mean per bone position error, or \emph{Bone Error}. It is similar to \emph{Joint Error}, but measures the \emph{relative} joint location accuracy. This metric is applicable for both 3D and 2D pose.

The next two are only for 3D pose as they measure the validity of 3D geometric constraints. Such metrics are important as violation of the constraints will cause physically infeasible 3D poses. Such errors are critical for certain applications such as 3D motion capture.

The second metric is the bone length standard deviation, or \emph{Bone Std}. It measures the stability of bone length. For each bone, the standard deviation of its length is computed over all the testing samples of the same subject.

The third metric is the percentage of illegal joint angle, or \emph{Illegal Angle}. It measures whether the rotation angles at a joint are physically feasible. We use the recent method and code in~\cite{akhter2015pose} to evaluate the legality of each predicted joint. Note that this metric is only for joints on the limbs and does not apply to those on the torso.

\begin{table}
\begin{center}
\begin{tabular}{lcccccccc}
\hline\noalign{\smallskip}
 & CNN prediction  & loss function\\
\noalign{\smallskip}
\hline
\noalign{\smallskip}
Baseline 		& joints $\mathcal{J}$ 		& $L(\mathcal{J})$, Eq.\eqref{eq.direct_joint_loss}	\\
\hline
Ours (joint) 	& \multirow{4}{*}{bones $\mathcal{B}$}  & $L(\mathcal{B},\mathcal{P}_{joint})$, Eq.\eqref{eq.compositional_loss_function}	\\
Ours (bone) 		&									    & 
$L(\mathcal{B},\mathcal{P}_{bone})$, Eq.\eqref{eq.compositional_loss_function}	\\
Ours (both) 		&								    	& 
$L(\mathcal{B},\mathcal{P}_{both})$, Eq.\eqref{eq.compositional_loss_function}	\\
Ours (all) 		&									    & 
$L(\mathcal{B},\mathcal{P}_{all})$, Eq.\eqref{eq.compositional_loss_function}	\\
\hline
\end{tabular}
\caption{The baseline and four variants of our method.}
\label{table:baselines}
\end{center}
\end{table}

\begin{table*}
\centering
\small
\begin{tabular}{l|c|c|c|c|c|c}
\hline
Training Data & Metric & Baseline & Ours (joint) & Ours(bone) & Ours (both) & Ours (all)\\
\hline \hline
\multirow{4}{*}{Human3.6M}
& Joint Error    & 102.2 & $103.3_{\uparrow1.1}  $ & $104.6_{\uparrow2.4}  $ 		   & $95.2_{\downarrow7.0}$  & $\textbf{92.4}_{\downarrow9.8}$  \\
& PA Joint Error & 75.0  & $74.3_{\downarrow0.7}$  & $75.0_{\downarrow0.0}$  		   & $68.1_{\downarrow6.9}$  & $\textbf{67.5}_{\downarrow7.5}$  \\
& Bone Error     & 65.5  & $63.5_{\downarrow2.0}$  & $62.3_{\downarrow3.2}$  		   & $59.1_{\downarrow6.4}$  & $\textbf{58.4}_{\downarrow7.1}$  \\
& Bone Std       & 26.4  & $23.9_{\downarrow2.5}$  & $21.9_{\downarrow4.5}$  		   & $22.3_{\downarrow4.1}$  & $\textbf{21.7}_{\downarrow4.7}$  \\
& Illegal Angle  & 3.7\% & $3.2\%_{\downarrow0.5}$ & $3.3\%_{\downarrow0.4}$  	   & $2.6\%_{\downarrow1.1}$ & $\textbf{2.5\%}_{\downarrow1.2}$ \\
\hline
\multirow{4}{*}{Human3.6M + MPII}
& Joint Error    & 64.2  & $62.9_{\downarrow1.3}$  & $63.8_{\downarrow0.4}$  		   & $60.7_{\downarrow3.5}$  & $\textbf{59.1}_{\downarrow5.1}$  \\
& PA Joint Error & 51.4  & $50.6_{\downarrow0.8}$  & $50.4_{\downarrow1.0}$  		   & $48.8_{\downarrow2.6}$  & $\textbf{48.3}_{\downarrow3.1}$  \\
&Bone Error  	 & 49.5  & $49.3_{\downarrow0.2}$  & $47.4_{\downarrow2.1}$		   & $47.2_{\downarrow2.3}$  & $\textbf{47.1}_{\downarrow2.4}$  \\
& Bone Std       & 19.9  & $19.3_{\downarrow0.6}$  & $\textbf{17.5}_{\downarrow2.4}$ & $17.6_{\downarrow2.3}$  & $18.0_{\downarrow1.9}$ 		  \\
\end{tabular}
\caption{Results of all methods under all evaluation metrics (the lower the better), with or without using MPII data in training. Note that the performance gain of all \emph{Ours} methods relative to the \emph{Baseline} method is shown in the subscript. The \emph{Illegal Angle} metric for ``Human3.6M+MPII'' setting is not included because it is very good ($<1\%$) for all methods.}
\label{table:hm36ablation}
\end{table*}

\begin{table*} 
\centering
\small
\begin{tabular}{l|c|c|c|c|c|c|c|c|c|c}
\hline
Metric &  \multicolumn{2}{|c|}{Joint Error} &  \multicolumn{2}{|c|}{PA Joint Error} 	&  \multicolumn{2}{|c|}{Bone Error} 	& \multicolumn{2}{|c|}{Bone Std} & \multicolumn{2}{|c|}{Illegal Angle}\\
\hline
Method & BL& Ours (all)& BL& Ours (all)& BL& Ours (all)& BL& Ours (all)& BL& Ours (all)\\
\hline \hline
Average    	    & 102.2 & $92.4_{\downarrow9.8}$   & 75.0  & $67.5_{\downarrow7.5}$   & 65.5  & $58.4_{\downarrow7.1}$  & 26.4 & $21.7_{\downarrow4.7}$   & 3.7\% & $2.5\%_{\downarrow1.2}$ \\
\hline
Ankle($\rightarrow$ Knee)   & 94.5  & $88.5_{\downarrow6.0}$   & 81.5  & $75.8_{\downarrow5.7}$   & 81.2  & $74.1_{\downarrow7.1}$  & 32.9 & $32.0_{\downarrow0.9}$   & - & - \\
Knee($\rightarrow$Hip)     & 68.6  & $63.7_{\downarrow4.9}$   & 69.2  & $62.9_{\downarrow6.3}$   & 69.1  & $63.4_{\downarrow5.7}$  & 21.7 & $22.8_{\uparrow1.1  }$   & 4.8\% & $3.8\%_{\downarrow1.0}$ \\
Hip($\rightarrow$Pelvis) 	    & 29.9  & $25.0_{\downarrow4.9}$   & 63.3  & $58.4_{\downarrow4.9}$   & 29.9  & $25.0_{\downarrow4.9}$  & 21.3 & $16.4_{\downarrow4.9}$   & 0.6\%     & $0.6\%_{\downarrow0.0}$     \\
Thorax($\rightarrow$Pelvis)   & 97.2  & $90.1_{\downarrow7.1}$   & 30.7  & $28.1_{\downarrow2.6}$   & 97.2  & $90.1_{\downarrow7.1}$  & 28.0 & $26.7_{\downarrow1.3}$   & -     & -     \\
Neck($\rightarrow$Thorax)     & 104.3 & $96.4_{\downarrow7.9}$   & 36.7  & $35.5_{\downarrow1.2}$   & 22.2  & $22.9_{\uparrow0.7  }$  & 12.4 & $11.7_{\downarrow0.7}$   & 2.2\%     & $1.3\%_{\downarrow0.9}$     \\
Head($\rightarrow$Neck)  & 115.4 & $108.4_{\downarrow7.0}$  & 42.8  & $41.1_{\downarrow1.7}$   & 39.7  & $37.3_{\downarrow2.4}$  & 15.3 & $14.8_{\downarrow0.5}$   & - & - \\
Wrist($\rightarrow$Elbow)     & 181.9 & $163.0_{\downarrow18.9}$ & 130.2 & $115.2_{\downarrow15.0}$ & 102.6 & $89.0_{\downarrow13.6}$ & 40.6 & $30.6_{\downarrow10.0}$  & - & - \\
Elbow($\rightarrow$Shoulder)  & 168.8 & $146.9_{\downarrow21.9}$ & 115.8 & $97.6_{\downarrow18.2}$  & 96.9  & $81.4_{\downarrow15.5}$ & 27.6 & $21.4_{\downarrow6.2}$   & 8.5\% & $5.4\%_{\downarrow3.1}$ \\
Shoulder($\rightarrow$Thorax) & 115.6 & $104.4_{\downarrow11.2}$ & 57.7  & $52.2_{\downarrow5.5}$   & 55.1  & $48.5_{\downarrow6.6}$  & 25.9 & $12.6_{\downarrow13.3}$  & 1.9\%     & $0.8\%_{\downarrow1.1}$     \\
\end{tabular}
\caption{Detailed results on all joints for \emph{Baseline} (BL) and \emph{Ours (all)} methods, only trained on Human3.6M data (top half in Table~\ref{table:hm36ablation}). The relative performance gain is shown in the subscript. Note that the left most column shows the names for both the joint (and the bone).}
\label{table:joint_error_hm36_only}
\end{table*}

\begin{figure*}[t]
\centering
\includegraphics [width=1.0\linewidth] {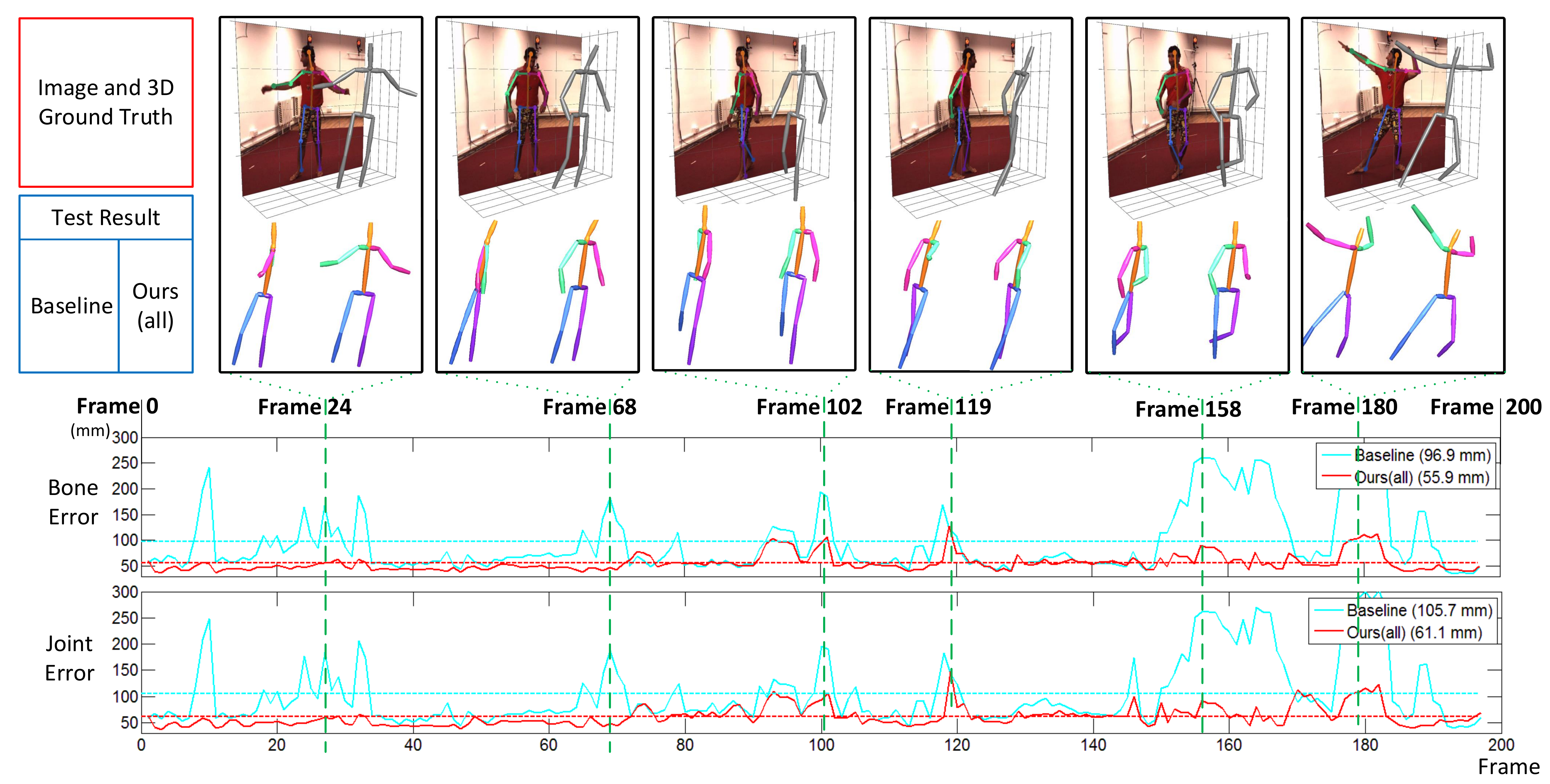}
\caption{(best viewed in color) Errors of wrist joint/bone of \emph{Baseline} and \emph{Ours (all)} methods on a video sequence from Human3.6M S9, action Pose. The average error over the sequence is shown in the legends. For this action, the arms have large motion and are challenging. Our method has much smaller joint and bone error. Our result is more stable over the sequence. The 3D predicted pose and ground truth pose are visualized for a few frames. More video results are at \url{https://www.youtube.com/watch?v=c-hgHqVK90M}.}
\label{fig.video_result}
\end{figure*}

\begin{table*}
\centering
\small
\begin{tabular}{l|c|c|c|c|c|c|c|c}
\hline
Method & Direction 	& Discuss & Eat &Greet 	& Phone & Pose 	& Purchase	& Sit	 \\
\hline \hline
Yasin\cite{yasin2016dual} & 88.4 & 72.5 & 108.5 & 110.2 & 97.1 & 81.6 & 107.2 & 119.0  \\
Rogez\cite{rogez2016mocap}& - & - & - & - & - & - & - & - \\
Chen\cite{chen20163d} & 71.6 & 66.6 & 74.7 & 79.1 & 70.1 & 67.6 & 89.3 & 90.7 \\
Moreno\cite{moreno20163d}& 67.4 & 63.8 & 87.2 & 73.9 & 71.5 & 69.9 & 65.1 & 71.7  \\
Zhou\cite{zhou2017monocap}& 47.9 & 48.8 & 52.7 & 55.0 & 56.8 & 49.0 & 45.5 & 60.8  \\
\hline
Baseline  & 45.2 & 46.0 & 47.8 & 48.4 & 54.6 & 43.8 & 47.0 & 60.6 \\
Ours(all)  & \textbf{42.1} & \textbf{44.3} & \textbf{45.0} & \textbf{45.4} & \textbf{51.5} & \textbf{43.2} & \textbf{41.3} & \textbf{59.3} \\
\hline \hline
Method & SitDown& Smoke 		& Photo   & Wait&Walk 	&WalkDog&WalkPair& Avg  \\
Yasin\cite{yasin2016dual} & 170.8 & 108.2 & 142.5 & 86.9 & 92.1 & 165.7 & 102.0 & 108.3 \\
Rogez\cite{rogez2016mocap}& - & - & - & - & - & - & - & 88.1 \\
Chen\cite{chen20163d} & 195.6& 83.5 & 93.3 & 71.2 & 55.7 & 85.9 & 62.5 & 82.7 \\
Moreno\cite{moreno20163d}& 98.6& 81.3 & 93.3 & 74.6 & 76.5 & 77.7 & 74.6 & 76.5  \\
Zhou\cite{zhou2017monocap}& 81.1& 53.7 & 65.5 & 51.6 & 50.4 & 54.8 & 55.9 & 55.3  \\
\hline
Baseline & 79.0& 54.5 & 56.0 & 46.7 & 42.2 & 51.0 & 47.9 & 51.4 \\
Ours (all)  & \textbf{73.3} & \textbf{51.0} & \textbf{53.0} & \textbf{44.0} & \textbf{38.3} & \textbf{48.0} & \textbf{44.8} & \textbf{48.3} \\
\end{tabular}
\caption{Comparison with previous work on Human3.6M. Protocol 1 is used. Evaluation metric is averaged \emph{PA Joint Error}. Extra 2D training data is used in all the methods. \emph{Baseline} and \emph{Ours (all)} use MPII data in the training. \emph{Ours (all)} is the best and also wins in all the $15$ activity categories.}
\label{table:hm36_p1_use_mpii}
\end{table*}

\begin{table*}
\centering
\small
\begin{tabular}{l|c|c|c|c|c|c|c|c}
\hline
Method & Direction 	& Discuss & Eat &Greet 	& Phone & Pose 	& Purchase	& Sit \\
\hline \hline
Chen\cite{chen20163d} & 89.9 & 97.6 & 90.0 & 107.9 & 107.3 & 93.6 & 136.1 & 133.1  \\
Tome\cite{tome2017lifting} & 65.0 & 73.5 & 76.8 & 86.4 & 86.3 & 68.9 & 74.8 & 110.2  \\
Moreno\cite{moreno20163d} & 69.5 & 80.2 & 78.2 & 87.0 & 100.8 & 76.0 & 69.7 & 104.7  \\
Zhou\cite{zhou2017monocap}& 68.7 & 74.8 & 67.8 & 76.4 & 76.3 & 84.0 & 70.2 & 88.0  \\
Jahangiri\cite{jahangiri2017generating}& 74.4 & 66.7 & 67.9 & 75.2 & 77.3 & 70.6 & 64.5 & 95.6  \\
Mehta\cite{mehta2016monocular}& 57.5 & 68.6 & 59.6 & 67.3 & 78.1 & 56.9 & 69.1 & 98.0  \\
Pavlakos\cite{pavlakos2016coarse}& 58.6 & 64.6 & 63.7 & 62.4 & 66.9  & 57.7 & 62.5 & 76.8 \\
\hline
Baseline  & 57.0 & 58.6 & 57.9 & 58.7 & 67.1 & 54.2 & 65.9 & 75.4  \\
Ours(all)  &\textbf{52.8} &\textbf{54.8} &\textbf{54.2} &\textbf{54.3} &\textbf{61.8} &\textbf{53.1} & \textbf{53.6} & \textbf{71.7}  \\
\hline \hline
Method & SitDown& Smoke 		& Photo   & Wait&Walk 	&WalkDog&WalkPair& Avg\\
\hline \hline
Chen\cite{chen20163d} & 240.1& 106.7 & 139.2 & 106.2 & 87.0 & 114.1 & 90.6 & 114.2 \\
Tome\cite{tome2017lifting} & 173.9& 85.0 & 110.7 & 85.8 & 71.4 & 86.3 & 73.1 & 88.4 \\
Moreno\cite{moreno20163d} & 113.9& 89.7 & 102.7 & 98.5 & 79.2 & 82.4 & 77.2 & 87.3 \\
Zhou\cite{zhou2017monocap}& 113.8& 78.0 & 98.4 & 90.1 & 62.6 & 75.1 & 73.6 & 79.9 \\
Jahangiri\cite{jahangiri2017generating}& 127.3& 79.6 & 79.1 & 73.4 & 67.4 & 71.8 & 72.8 & 77.6 \\
Mehta\cite{mehta2016monocular}& 117.5&69.5  & 82.4 & 68.0 & 55.3 & 76.5 & 61.4 & 72.9 \\
Pavlakos\cite{pavlakos2016coarse}& 103.5& 65.7 & 70.7 & 61.6 & 56.4 & 69.0 & 59.5 & 66.9 \\
\hline
Baseline & 98.1& 66.2 & 71.1 & 58.4 & 51.4 & 65.2 & 56.8 & 64.2 \\
Ours(all)& \textbf{86.7}& \textbf{61.5} & \textbf{67.2} & \textbf{53.4} & \textbf{47.1} & \textbf{61.6} & \textbf{53.4} & \textbf{59.1}\\
\end{tabular}
\caption{Comparison with previous work on Human3.6M. Protocol 2 is used. Evaluation metric is averaged \emph{Joint Error}. Extra 2D training data is used in all the methods. \emph{Baseline} and \emph{Ours (all)} use MPII data in the training. \emph{Ours (all)} is the best and also wins in all the $15$ activity categories.}
\label{table:hm36_p2_use_mpii}
\end{table*}

\begin{table*}
\centering
\small
\begin{tabular}{l|c|c|c|c|c|c|c|c}
\hline
Method & Direction 	& Discuss & Eat &Greet 	& Phone & Pose 	& Purchase	& Sit \\
\hline \hline
Zhou\cite{zhou2016sparseness} & \textbf{87.4} & 109.3 & 87.1	& 103.2 &116.2 &106.9 &99.8 &124.5 \\
Tekin\cite{tekin2016direct}   & 102.4 & 147.7 & 88.8	& 125.4 &118.0 &112.4 &129.2 &138.9 \\
Xingyi\cite{zhou2016deep}     & 91.8 & 102.4 & 97.0 & 98.8 & 113.4 &90.0 & 93.8 &132.2  \\
\hline
Baseline	                  & 98.8 & 101.9 & 89.8 &89.9 &100.0 &97.2 &113.2 & 102.0 \\
Ours(all)		              & 90.2 & \textbf{95.5} & \textbf{82.3} &\textbf{85.0} &\textbf{87.1} & \textbf{87.9} & \textbf{93.4} &\textbf{100.3} \\
\hline \hline
Method & SitDown & Smoke 		& Photo   & Wait&Walk 	&WalkDog&WalkPair& Avg  \\
\hline \hline
Zhou\cite{zhou2016sparseness} &199.2& 107.4 & 139.5 &118.1 &79.4 &114.2 &97.7 &113.0 \\
Tekin\cite{tekin2016direct}   & 224.9& 118.4 & 182.7&138.8 &\textbf{55.1} &126.3 &\textbf{65.8} &125.0 \\
Xingyi\cite{zhou2016deep}     & 159.0& 106.9 & 125.2 &94.4 &79.0 &126.0 &99.0 &107.3 \\
\hline
Baseline					  & 138.9& 101.7 & 101.1 & 95.9 & 90.8 & 108.9 & 102.7 & 102.2 \\
Ours(all)					  & \textbf{135.4}& \textbf{91.4} & \textbf{94.5} &\textbf{87.3}	&78.0 &\textbf{90.4} &86.5 &\textbf{92.4}\\
\end{tabular}
\caption{Comparison with previous work on Human3.6M. Protocol 2 is used. Evaluation metric is averaged \emph{Joint Error}. No extra training data is used. \emph{Ours (all)} is the best and wins in $12$ out of $15$ activity categories. Note that Tekin et al. \cite{tekin2016direct} report more accurate results for "Walk" and "WalkPair", but their method uses the temporal context information in the video. Our method only runs on individual frames.}
\label{table:hm36_p2_no_mpii}
\end{table*}

\begin{figure*}[t]
\centering
\includegraphics [width=1.0\linewidth] {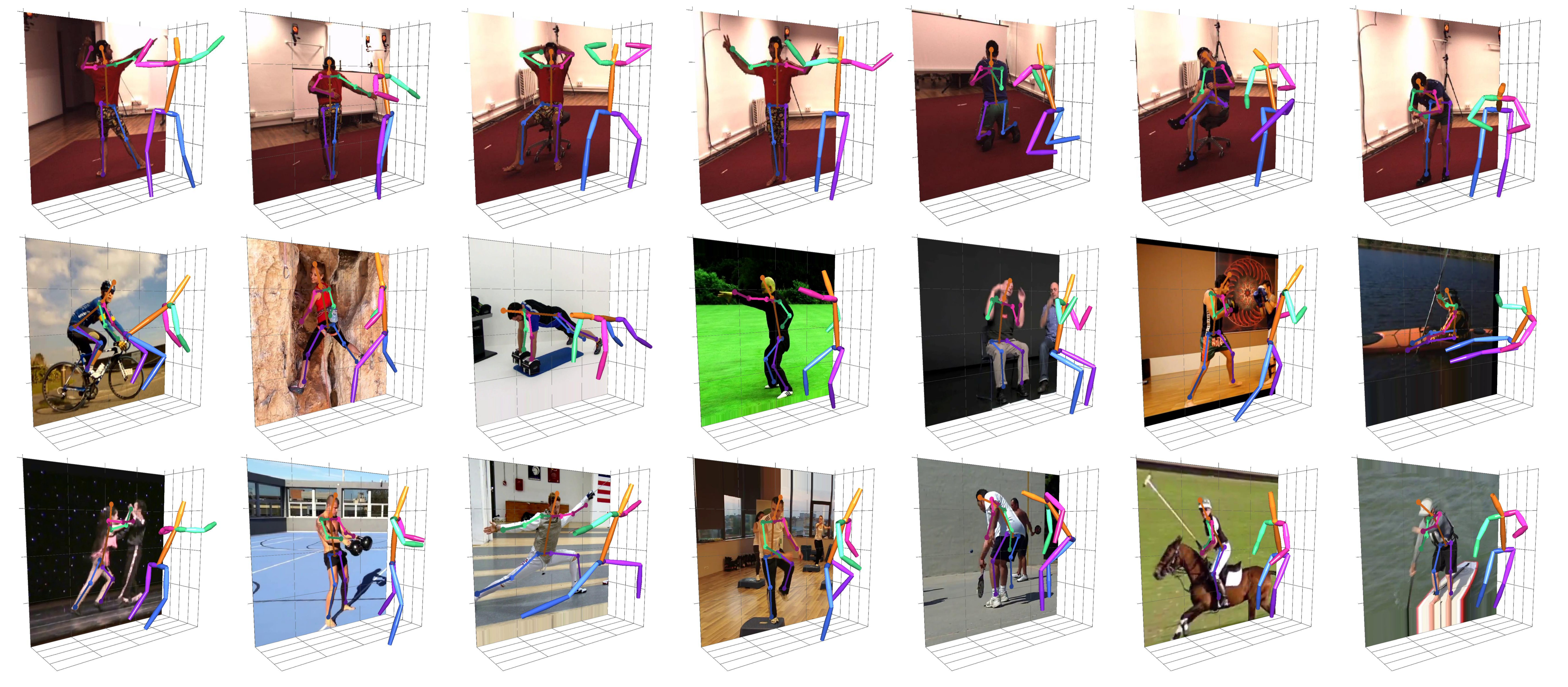}
\caption{(best viewed in color) Examples of 3D pose estimation for Human3.6M (top row) and MPII (middle and bottom rows), using \emph{Ours (all)} method in Table~\ref{table:hm36_p2_use_mpii}, trained with both 3D and 2D data. Note that the MPII 3D results are quite plausible and convincing.}
\label{fig.qualitative_result}
\end{figure*}

\begin{table}[t]
\centering
\footnotesize
\begin{tabular}{c|c|c|c|c|c|c}
\hline
Stage & Metric & IEF* & joint & bone & both & all\\
\hline \hline
\multirow{3}{*}{ 0}
& Joint Error   &29.7   &27.2   &27.8   			&27.5 	&\textbf{27.2}  \\
& Bone Error 	&24.8   &23.1   &\textbf{22.1}   	&22.7 	&22.5  \\
& PCKH 0.5 		&76.5\% &79.3\% &79.0\% 			&79.2\%	&\textbf{79.6\%}\\
\hline \hline
\multirow{3}{*}{ 1}
& Joint Error   &25.0   &23.8   &25.2   &23.0    &\textbf{22.8}   \\
& Bone Error	&21.2   &20.5   &20.9   &19.7    &\textbf{19.5}   \\
& PCKH 0.5 		&82.9\% &84.1\% &82.7\% &84.9\%  &\textbf{86.4\%} \\
\end{tabular}
\caption{Results of the baseline and four variants of our method (see Table~\ref{table:baselines}), in the two-stage IEF*.}
\label{table:mpii_ablation}
\end{table}

\begin{table}
\centering
\footnotesize
\addtolength{\tabcolsep}{-2.5pt}
\begin{tabular}{l|c|c|c|c|c|c|c|c}
\hline
Method & Head & Sho. & Elb. &Wri. &Hip&Knee&Ank.&Mean\\
\hline \hline
Pishchulin \cite{pishchulin2013strong}&74.3 &49.0 &40.8 &34.1 &36.5 &34.4 &35.2 &44.1\\
Tompson\cite{tompson2014joint}		  &95.8 &90.3 &80.5 &74.3 &77.6 &69.7 &62.8 &79.6\\
Tompson\cite{tompson2015efficient}	  &96.1 &91.9 &83.9 &77.8 &80.9 &72.3 &64.8 &82.0\\
Hu\cite{hu2016bottom}				  &95.0 &91.6 &83.0 &76.6 &81.9 &74.5 &69.5 &82.4\\
Pishchulin\cite{pishchulin2016deepcut}&94.1 &90.2 &83.4 &77.3 &82.6 &75.7 &68.6 &82.4\\
Lifshitz\cite{lifshitz2016human}	  &97.8 &93.3 &85.7 &80.4 &85.3 &76.6 &70.2 &85.0\\
Gkioxary\cite{gkioxari2016chained}	  &96.2 &93.1 &86.7 &82.1 &85.2 &81.4 &74.1 &86.1\\
Raf\cite{rafi2016efficient}			  &97.2 &93.9 &86.4 &81.3 &86.8 &80.6 &73.4 &86.3\\
Insafutdinov\cite{insafutdinov2016deepercut} &96.8 &95.2 &89.3 &84.4 &88.4 &83.4 &78.0 &88.5\\
Wei\cite{wei2016convolutional}		  &97.8 &95.0 &88.7 &84.0 &88.4 &82.8 &79.4 &88.5\\
Bulat\cite{bulat2016human}			  &97.9 &95.1 &89.9 &85.3 &89.4 &85.7 &81.7 &89.7\\
Newell\cite{newell2016stacked}		  &98.2 &96.3 &91.2 &87.1 &90.1 &87.4 &83.6 &90.0\\
Chu\cite{chu2017multi}				  &\textbf{98.5} &\textbf{96.3} &\textbf{91.9} &\textbf{88.1} &\textbf{90.6} &\textbf{88.0} &\textbf{85.0} &\textbf{91.5}\\
\hline \hline
Carreira(IEF)\cite{carreira2016human} &95.7 &91.7 &81.7 &72.4 &82.8 &73.2 &66.4 &81.3\\
IEF*					  		 	  &96.3 &92.6 &83.1 &74.6 &83.7 &74.1 &71.4 &82.9\\
Ours (all) 							  &\textbf{97.5} &\textbf{94.3} &\textbf{87.0} &\textbf{81.2} &\textbf{86.5} &\textbf{78.5} &\textbf{75.4} &\textbf{86.4}\\
\end{tabular}
\addtolength{\tabcolsep}{2.5pt}
\caption{Comparison to state-of-the-art works on MPII (top: detection based, bottom: regression based). PCKH 0.5 metric is used. Our approach significantly improves the baseline IEF and is competitive to other detection based methods.}
\label{table:mpii_comparison}
\end{table}

\subsection{Experiments on 3D Pose of Human3.6M}
For Human3.6M~\cite{ionescu2014human3}, there are two widely used evaluation protocols with different training and testing data split.

\emph{Protocol 1} Six subjects (S1, S5, S6, S7, S8, S9) are used in training. Evaluation is performed on every 64th frame of Subject 11's videos. It is used in~\cite{yasin2016dual,rogez2016mocap,chen20163d,moreno20163d,zhou2017monocap}. \emph{PA Joint Error} is used for evaluation.


\emph{Protocol 2} Five subjects (S1, S5, S6, S7, S8) are used for training. Evaluation is performed on every 64th frame of two subjects (S9, S11). It is used in~\cite{zhou2016sparseness,tekin2016direct,zhou2016deep,chen20163d,tome2017lifting,moreno20163d,zhou2017monocap,jahangiri2017generating,mehta2016monocular,pavlakos2016coarse}. \emph{Joint Error} is used for evaluation.

\textbf{Ablation study}. The \emph{direct joint regression} baseline and four variants of our method are compared. They are briefly summarized in Table \ref{table:baselines}.
As explained in Section~\ref{sec.2d_3d_train}, training can use additional 2D data (from MPII), optionally. We therefore tested two sets of training data: 1) only Human3.6M; 2) Human3.6M plus MPII.

Table~\ref{table:hm36ablation} reports the results under Protocol 2, which is more commonly used. We observe several conclusions. 

\emph{Using 2D data is effective.} All metrics are significantly improved after using MPII data. For example, joint error is reduced from $102.2$ to $64.2$. This improvement should originate from the better learnt feature from the abundant 2D data. See the contemporary work~\cite{zhou2016towards} for more discussions. Note that adding 2D data in this work is simple and not considered as a main contribution. Rather, it is considered as a baseline to validate our regression approach.

\emph{Bone representation is superior than joint representation.} This can be observed by comparing \emph{Baseline} with \emph{Ours (joint)} and \emph{Ours (bone)}. They are comparable because they use roughly the same amount of supervision signals in the training. The two variants of ours are better on nearly all the metrics, especially the geometric constraint based ones. 


\emph{Compositional loss is effective.} When the loss function becomes better (\emph{Ours (both)} and \emph{Ours (all)}), further improvement is observed. Specifically, when trained only on Human3.6M, \emph{Ours (all)} improves the Baseline by $9.8$ mm (relative $9.6\%$) on joint error, $7.5$ mm (relative $10\%$) on PA joint error, $7.1$ mm (relative $10.8\%$) on bone error, $4.7$ mm (relative $17.8\%$) on bone std, and $1.2\%$ (relative $32.4\%$) on illegal angle.

Table \ref{table:joint_error_hm36_only} further reports the performance improvement from \emph{Ours (all)} to \emph{Baseline} on all the joints (bones). It shows several conclusions. First, \emph{limb joints are harder than torso joints and upper limbs are harder than lower limbs}. This is consistent as Figure~\ref{fig.bone_representation} (middle). It indicates that the variance is a good indicator of difficulty and a per-joint analysis is helpful in both algorithm design and evaluation. Second, \emph{our method significantly improves the accuracy for all the joints}, especially the challenging ones like wrist, elbow and ankle. Figure~\ref{fig.video_result} shows the results on a testing video sequence with challenging arm motions. Our result is much better and more stable.

\textbf{Comparison with the state-of-the-art}
There are abundant previous works. They have different experiment settings and fall into three categories. They are compared to our method in Table~\ref{table:hm36_p1_use_mpii}, \ref{table:hm36_p2_use_mpii}, and~\ref{table:hm36_p2_no_mpii}, respectively. 


The comparison is not completely fair due to the differences in the training data (when extra data are used), the network architecture and implementation. Nevertheless, two common conclusions validate that \emph{our approach is effective and sets the new state-of-the-art in all settings by a large margin}. First, \emph{our baseline is strong}. It is simple but already improves the state-of-the-art, by $3.9$ mm (relative $7\%$) in Table~\ref{table:hm36_p1_use_mpii}, $2.7$ mm (relative $4\%$) in Table~\ref{table:hm36_p2_use_mpii}, and $5.1$ mm (relative $4.8\%$) in Table~\ref{table:hm36_p2_no_mpii}. Therefore, it serves as a competitive reference. Second, \emph{our method significantly improves the baseline, using exactly the same network and training}. Thus, the improvement comes from the new pose representation and loss function. It improves the state-of-the-art significantly, by $7$ mm (relative $12.7\%$) in Table~\ref{table:hm36_p1_use_mpii}, $7.8$ mm (relative $11.7\%$) in Table~\ref{table:hm36_p2_use_mpii}, and $14.9$ mm (relative $13.9\%$) in Table~\ref{table:hm36_p2_no_mpii}. 

Example 3D pose results are illustrated in Figure~\ref{fig.qualitative_result}.


\subsection{Experiments on 2D Pose of MPII}

All leading methods on MPII benchmark~\cite{mpiiwebpage} have sophisticated network architectures. As discussed in Section~\ref{sec.related_work}, the best-performing family of methods adopts a multi-stage architecture~\cite{chu2017multi, carreira2016human, newell2016stacked, bulat2016human, wei2016convolutional, insafutdinov2016deepercut, gkioxari2016chained}. Our method is novel in the pose representation and loss function. It is complementary to such sophisticated networks. In this experiment, it is integrated into the \emph{Iterative Error Feedback} method (IEF)~\cite{carreira2016human}, which is the only regression based method in the family.

We implement a two stage baseline IEF, using ResNet-50 as the basic network in each stage. For reference, the original IEF~\cite{carreira2016human} uses five stages with GoogLeNet for each stage. We denote our implementation as \emph{IEF*}. The two stages in \emph{IEF*} are then modified to use our bone based representation and compositional loss function. The training for all the settings remains the same, as specified in Section~\ref{sec.2d_3d_train}.

\textbf{Ablation study} Table~\ref{table:mpii_ablation} shows the results of IEF* and our four variants. We observe the same conclusions as in Table~\ref{table:hm36ablation}. Both bone based representation and compositional loss function are effective under all metrics. In addition, both stages in IEF* benefit from our approach.

\textbf{Comparison with the state-of-the-art} Table~\ref{table:mpii_comparison} reports the comparison result to state-of-the-art works on MPII. PCKH0.5 metric is used. Top section of Table~\ref{table:mpii_comparison} is detection based methods and bottom section is regression based. Ours (86.4\%) produces significant improvement over the baseline (IEF*) and becomes the best regression based method. It is competitive to other detection based methods.

\section{Conclusion}
We show that regression based approach is competitive to the leading detection based approaches for 2D pose estimation once pose structure is appropriately exploited. Our approach is more potential for 3D pose estimation, where more complex structure constraints are critical.

\section*{Acknowledgement}
This research work was supported by The National Science Foundation of China No. 61305091, and the Fundamental Research Funds for the Central Universities No. 2100219054.

{\small
\bibliographystyle{ieee}
\bibliography{egbib}
}

\end{document}